%% file: main.tex
\documentclass[twoside]{article}

\usepackage{graphicx}
\usepackage{tikz}
\usepackage{url}
\usetikzlibrary{shapes,decorations,arrows,calc,arrows.meta,fit,positioning}
\tikzset{
    -Latex,auto,node distance =1 cm and 1 cm,semithick,
    state/.style ={ellipse, draw, minimum width = 0.7 cm},
    point/.style = {circle, draw, inner sep=0.04cm,fill,node contents={}},
    bidirected/.style={Latex-Latex,dashed},
    el/.style = {inner sep=2pt, align=left, sloped}
}
\usepackage{amsmath}
\usepackage[accepted]{aistats2026}

\usepackage[round]{natbib}

\begin{document}

\twocolumn[

\aistatstitle{The Cross-Context Threshold Test: Detecting Discrimination Under Environmental Shifts}

\aistatsauthor{Jun Yuan \And Xinyue Ye}

\aistatsaddress{\
Digital Technology for Democracy Lab \\ 
University of Virginia \\
\texttt{uzz3jj@virginia.edu}
\And
Department of Geography and the Environment \\
University of Alabama \\
\texttt{xye10@ua.edu} 
} ]

\begin{abstract}
  We study how threshold tests for detecting discrimination under environmental shifts, focusing on the Veil-of-Darkness~(VoD) setting where visibility changes between daylight and darkness. We show that standard threshold tests, when applied separately to daylight and darkness data, violate key assumptions: risk distributions drift across contexts and thresholds fluctuate arbitrarily. We propose a cross-context threshold test that enforces distributional invariance and monotonic threshold decay. Using New York City stop-and-frisk data and synthetic experiments, we demonstrate that this model yields more reliable thresholds, improves bias detection, and aligns with the counterfactual logic of the VoD test. Our framework generalizes to fairness auditing whenever environmental context influences decisions.
\end{abstract}

\input{sections/intro}

\input{sections/relatedWork}

\input{sections/method}

\input{sections/caseStudy}

\input{sections/Discussion}

\input{sections/conclusion}

\subsubsection*{Acknowledgements}
We thank the anonymous reviewers for their valuable feedback. We also thank members of the Digital Technology for Democracy Lab at the University of Virginia and Xinyue Ye’s research group for helpful discussions and suggestions. This work was supported in part by the University of Virginia.

\bibliographystyle{plainnat}
\bibliography{ref}  

\section*{Checklist}

\begin{enumerate}

  \item For all models and algorithms presented, check if you include:
  \begin{enumerate}
    \item A clear description of the mathematical setting, assumptions, algorithm, and/or model. [Yes]
    \item An analysis of the properties and complexity (time, space, sample size) of any algorithm. [No]
    \item (Optional) Anonymized source code, with specification of all dependencies, including external libraries. [Yes]
  \end{enumerate}

  \item For any theoretical claim, check if you include:
  \begin{enumerate}
    \item Statements of the full set of assumptions of all theoretical results. [Yes]
    \item Complete proofs of all theoretical results. [Yes]
    \item Clear explanations of any assumptions. [Yes]     
  \end{enumerate}

  \item For all figures and tables that present empirical results, check if you include:
  \begin{enumerate}
    \item The code, data, and instructions needed to reproduce the main experimental results (either in the supplemental material or as a URL). [Yes]
    \item All the training details (e.g., data splits, hyperparameters, how they were chosen). [Yes]
    \item A clear definition of the specific measure or statistics and error bars (e.g., with respect to the random seed after running experiments multiple times). [Yes]
    \item A description of the computing infrastructure used. (e.g., type of GPUs, internal cluster, or cloud provider). [Yes]
  \end{enumerate}

  \item If you are using existing assets (e.g., code, data, models) or curating/releasing new assets, check if you include:
  \begin{enumerate}
    \item Citations of the creator If your work uses existing assets. [Yes]
    \item The license information of the assets, if applicable. [Yes]
    \item New assets either in the supplemental material or as a URL, if applicable. [Yes]
    \item Information about consent from data providers/curators. [Not Applicable]
    \item Discussion of sensible content if applicable, e.g., personally identifiable information or offensive content. [Not Applicable]
  \end{enumerate}

  \item If you used crowdsourcing or conducted research with human subjects, check if you include:
  \begin{enumerate}
    \item The full text of instructions given to participants and screenshots. [Not Applicable]
    \item Descriptions of potential participant risks, with links to Institutional Review Board (IRB) approvals if applicable. [Not Applicable]
    \item The estimated hourly wage paid to participants and the total amount spent on participant compensation. [Not Applicable]
  \end{enumerate}

\end{enumerate}

\clearpage
\appendix
\thispagestyle{empty}

\onecolumn
\aistatstitle{The Cross-Context Threshold Test: Detecting Discrimination Under Environmental Shifts:
Supplementary Materials}

\section{Proof in Section 3.2 Assumption 2}

We aim to show that thresholds across races exhibit reduced variance at night compared to daytime.

Let the daytime thresholds be
\[
t = (t_1, t_2, \dots, t_n)^\top \in [0,1]^n,
\]
and the corresponding nighttime thresholds be
\[
t' = (t'_1, t'_2, \dots, t'_n)^\top,
\]
where for each race \( i \),
\[
t'_i = a_i t_i, \quad a_i > 0.
\]
Here \( a_i \) represents the ratio of the nighttime threshold to the daytime threshold for race \( i \).

We assume that nighttime thresholds are not inflated overall relative to daytime thresholds:
\begin{equation}
\label{eq:bounded_norm}
\|t'\| \le \|t\|, 
\quad \text{or equivalently,} \quad
0 \le \frac{\|t'\|}{\|t\|} \le 1.
\end{equation}

To quantify cross-racial disparity, we define the pairwise variance functional
\[
V(t) = \sum_{i<j} (t_i - t_j)^2 .
\]
The corresponding nighttime variance is
\[
V(t') = \sum_{i<j} (t'_i - t'_j)^2 
= \sum_{i<j} (a_i t_i - a_j t_j)^2 .
\]
The ratio between night and day variances is then
\begin{equation}
\label{eq:ratio_def}
R = \frac{V(t')}{V(t)} 
= \frac{\sum_{i<j} (a_i t_i - a_j t_j)^2}{\sum_{i<j} (t_i - t_j)^2}.
\end{equation}

Each term in the numerator of~\eqref{eq:ratio_def} scales the daytime difference 
\((t_i - t_j)^2\) by a factor that depends smoothly on \(a_i\) and \(a_j\).
Since all weights are positive and the function is quadratic in the \(a_i\)’s,
the overall ratio \(R\) must lie between the smallest and largest squared scaling factors:
\begin{equation}
\label{eq:variance_bound_simple}
\boxed{
\min_i a_i^2
\;\le\;
R
\;\le\;
\max_i a_i^2 .
}
\end{equation}

If \( 0 < a_i < 1 \) for all \( i \), then both bounds in~\eqref{eq:variance_bound_simple} 
are strictly less than one, implying
\[
V(t') < V(t).
\]
Therefore, the cross-racial threshold variance decreases at night, 
indicating that nighttime thresholds are more homogeneous across races.

\end{document}

%% file: sections/intro.tex
\section{INTRODUCTION}

Auditors tasked with detecting bias in decision-making systems often face two challenges:  
(1) adapting existing statistical tests to new contexts, and  
(2) developing tools that make audit outcomes interpretable and actionable.  
Both remain under-explored.  
In policing, for example, officers stop cars or pedestrians to search for contraband.  
The underlying decision process can be viewed as a binary classifier: each individual has a latent probability of carrying contraband, and an officer chooses a threshold above which to initiate a stop.  
Observed stop outcomes (successful versus unsuccessful stops) provide indirect evidence of these thresholds.  
A recent work introduced the \emph{threshold test}, which estimates decision thresholds for racial groups from observed stop counts, hit counts, and census-based population proportions~\citep{pierson2018fast} .  
By jointly estimating both group-level risk distributions and thresholds, the threshold test addresses the well-known problem of inframarginality \citep{arnold2018racial, ayres2002outcome, engel2008critique, pierson2020large}.  
A complementary strategy is the \emph{veil-of-darkness} (VoD) test \citep{grogger2006testing, worden2012testing, knode2024pulling}, which exploits a natural experiment: after sunset, reduced visibility makes it more difficult for officers to identify a person’s race before deciding to stop them.  
The VoD test assumes that underlying risk distributions remain stable across twilight, so that any observed changes in stop rates or thresholds can be attributed to racial bias.  
Recent work often combines threshold and VoD tests to study racial disparities in policing \citep{pierson2020large, knode2024pulling}.  
Some researchers extend this by applying threshold tests separately to daylight and darkness subsets, interpreting differences in estimated thresholds as evidence aligned with VoD logic—that racial disparities diminish when race is visually obscured \citep{pierson2020large}.  
However, this direct application overlooks key VoD assumptions: (1) the risk distribution for any racial group should not shift abruptly with visibility, and (2) thresholds should converge under reduced visibility.  
When these assumptions are violated, the test results are unstable or misleading.  

To address this gap, we propose the \emph{cross-context threshold test} (CCTT), which adapts the threshold test to explicitly account for VoD conditions.  
Our model enforces both distributional invariance across daylight and darkness and monotonic threshold decay under reduced visibility.  
Through controlled simulation studies, we demonstrate that CCTT more accurately recovers ground-truth thresholds and achieves higher recall in detecting bias than the standard threshold test.  
Applied to New York City stop-and-frisk data, CCTT uncovers additional racial disparities and better aligns with the counterfactual logic of the VoD framework.  
Beyond policing, this approach generalizes to other high-stakes domains where context shifts alter decision thresholds but not underlying risk distributions.

%% file: sections/relatedWork.tex
\section{BACKGROUND}

\subsection{Threshold Tests for Discrimination}
Outcome tests compare success rates (e.g., hit rates) across groups but suffer from the inframarginality problem: differences in hit rates do not necessarily imply differences in decision thresholds \citep{ayres2002outcome, engel2008critique, simoiu2017problem}. Threshold tests address this limitation by jointly estimating group-level risk distributions and the thresholds applied to those distributions \citep{pierson2018fast}. The model infers how officers set stop thresholds across racial groups using observed stop counts, hit counts, and population proportions. This approach has become a leading method for identifying disparate treatment in policing data and has been applied at scale across jurisdictions in the United States \citep{pierson2020large}.

\subsection{The Veil-of-Darkness (VoD) Framework}
The veil-of-darkness test, introduced by~\cite{grogger2006testing}, provides a complementary quasi-experimental approach. It exploits the natural transition around sunset: after dark, reduced visibility makes it more difficult for officers to discern a pedestrian’s or driver’s race before deciding to stop them. The core assumption is that the underlying risk distribution of contraband possession does not change abruptly around twilight. Thus, any observed changes in stop rates or thresholds across daylight and darkness can be interpreted as evidence of bias in policing practices \citep{knode2024pulling, worden2012testing}.

\subsection{The Gap}
Recent work~\citep{pierson2020large} has attempted to combine these approaches by applying threshold tests separately to daylight and darkness stops, interpreting threshold differences as evidence aligned with VoD logic. However, this direct application violates key assumptions. Risk distributions are estimated independently across day and night, leading to instability, and thresholds may fluctuate arbitrarily. As we show in Section~\ref{sec:dist_violate}, this motivates the \emph{cross-context threshold test} (CCTT), which explicitly enforces distributional invariance and monotonic threshold decay across visibility conditions.

%% file: sections/method.tex
\section{THE CROSS-CONTEXT THRESHOLD TEST (CCTT)}

To analyze how environmental context influences bias detection, we formalize three components of our framework: (1) data curation under contextual conditions, (2) the classification task being audited, and (3) the bias testing method. This structure allows us to clearly identify where context enters the analysis and how it shapes the interpretation of bias metrics.

\subsection{{Problem Setup}}

Let $\mathcal{D} = \{\mathbf{x}_i, y_i\}_{i=1}^N$ be a dataset of $N$ decision instances, where $\mathbf{x}_i$ is the feature vector and $y_i$ is the observed decision outcome (e.g., whether a stop yielded contraband). In many applications, we analyze a curated subset of $\mathcal{D}$ that meets specific contextual conditions. In our case, the veil-of-darkness (VoD) setting. We denote this curated dataset as:
\[
\mathcal{D}_K = f(\mathcal{D}, K)
\]
where $K$ encodes two kinds of auxiliary conditions: 
\begin{enumerate}
    \item \textbf{Bias-analysis scope} ($K_{\text{bias}}$): the subset of instances and outcomes under analysis (e.g., a specific region, time window, or racial groups of interest).
    \item \textbf{Distribution priors} ($K_{\text{dist}}$): assumptions about underlying population composition or distributions.
\end{enumerate}
If no auxiliary conditions are applied, $\mathcal{D}_K = \mathcal{D}$.

The decision process being audited is represented as a function $h$ mapping features to outcomes:
\[
h = \mathcal{M}(\mathcal{D}_K)
\]
Here, $h$ may be as simple as a lookup from the dataset (when outcomes are directly observed) or a model that generalizes to new instances from a training paradigm $\mathcal{M}$. In this work, we treat $h$ as a lookup function to focus on auditing observed historical decisions.

A \textbf{bias test} computes a group-level metric $\mathbf{t} = \{t_{p_1}, \dots, t_{p_s}\}$ for each of $s$ groups $\mathbf{p} = \{p_i\}_{i=1}^s$:
\[
\mathbf{t} = \mathcal{B}(h, \mathcal{D}_K)
\]
Examples of $t_{p_i}$ include positive rates (PR), true positive rates (TPR), or decision thresholds. A statistically significant difference among these values suggests potential bias. When the bias test is adapted to a specific context $K$, we write $\mathcal{B}_K$.

\subsection{Assumptions}

In this section, we describe the assumptions $K$ for the VoD threshold test to define the $\mathcal{B}_K$.

\textbf{Assumption 1}: Risk~(signal) distribution for each racial group does not change from daylight to darkness during the intertwilight period.

\textbf{Assumption 2}: Police always find it more difficult to distinguish race at darkness than daylight due to limited visibility.

The Assumption 1 reflects a simple intuition that while group risk distributions may evolve over long time period~(e.g., years of demographic or policy change), it is reasonable to treat them as stable across the short temporal window of twilight.
This is consistent with the VoD literature~\citep{worden2012testing}, which recommends restricting analysis to a narrow twilight window where population characteristics are held constant except for visibility.
The implication of Assumption 2 may be interpreted as the thresholds are more similar~(i.e., smaller variance) at darkness than daylight among racial groups.
Direct enforcement of this assumption is difficult.  
As a practical alternative, we impose that the threshold in darkness is less than or equal to its daylight counterpart.
This condition reflects the interpretation that officers exercise a less selective~(i.e., more cautious) decision rule at night for the same racial group. 
Such behavior is consistent with the VoD literature~\citep{grogger2006testing} which suggests that reduced visibility leads officers to rely on lower stopping thresholds.
When thresholds are restricted to the interval $[0,1]$, this condition also implies Assumption 2~(proof provided in the Supplementary Material).
Importantly, lower thresholds do not necessarily translate into a higher number of stops, as stop counts also depend on factors such as street population and patrol exposure.

\input{sections/simulationStudy}

\subsection{Logit-Normal Mixture Model}
\label{sec:cctt}
In this section, we describe modifications to the threshold test models introduced in prior work.  
The original threshold test represents group-level risk distributions using Beta distributions \citep{simoiu2017problem}.  
Because the Beta distribution is naturally supported on the interval $[0,1]$, this formulation also constrains estimated thresholds to lie within that range.
Bayesian estimation for beta distributions is time-consuming. 
A fast threshold test is proposed to utilize a logit-normal mixture model as a proxy for the Beta distribution, thereby speeding up the test by approximately 40 times while achieving the same test results~\citep{pierson2018fast}.
They apply the threshold test to the NYC stop-and-frisk dataset.
Each instance in the dataset represents a pedestrian stop record at a certain time and location in New York City.  Police search~(frisk) decision is recorded~(searched or not searched), as well as the outcome~(a hit or not a hit).
A hit refers to the fact that the police are correct in their search decisions~(i.e., the discovery of contraband).
They also propose a threshold test for search~(frisk) decision and a threshold test for stop decision by estimating the racial proportion of pedestrians in each police precinct using Census data.

We focus on the threshold test for \emph{stop decisions}, as opposed to searches, because they are more directly influenced by visibility: an officer may not discern a pedestrian’s race at a distance in darkness, but will almost always do so before deciding to frisk.

Here, we introduce the complete fast threshold test.
$Y$ is the result of stop (guilty = 1, innocent = 0) based on the risk probability $z$.
\begin{equation} 
Y = Bernoulli(z)
\end{equation}

$X$ is the risk signal of the person~(pedestrian) that follows a normal distribution given that they are guilty or innocent.
\begin{equation} 
X|Y=0 \sim N(\mu_ 0, \sigma_0)
\end{equation}

\begin{equation} 
X|Y=1 \sim N(\mu_ 1, \sigma_1)
\end{equation}

\begin{equation}
   Z = g(x; \mu_ 0, \sigma_ 0, \mu_ 1, \sigma_1) =  P(Y=1 | X=x)
\end{equation}

$g(x)$ is the risk function that maps a real value $x$ to the probability of the risk bounded between 0 and 1, denoted as $Z$. 
Determining the risk distribution for each race is achieved by estimating $\mu_ 0$, $\sigma_ 0$, $\mu_ 1$, $\sigma_ 1$ per race.
We can reduce the number of parameters to be estimated following prior work~\citep{pierson2018fast}.To ensure heteroskedasticity for $g$, we constrain $\sigma_0 = \sigma_1$. This homoskedasticity condition ensures that a higher signal $X$ corresponds to a higher risk probability $Z$ (i.e., monotonicity); otherwise, the likelihood may become non-monotonic and thresholds may be non-identifiable.  Without loss of generality, we set $\mu_0 = 0$ and estimate only $\mu_1$. Setting $\mu_0 = 0$ for each group is a standard location normalization that does not affect cross-group comparability. Only the difference between $\mu_0$ and $\mu_1$ determines the mapping from $X$ to risk $Z$, and thresholds are estimated in $Z$-space, which is invariant to affine shifts.
While estimating the distribution, we also need to estimate the stop threshold $t$ per race per police region simultaneously. 
We fit the model to the ground truth data, such as hit count and stop count, to determine the best model parameter values. 

\textit{Step 1: For each race, we fit the stop count and hit count per police region}:

Let $S$ be the stop count. 
\begin{equation}
    S \sim Binomial( \text{population}, \text{stop rate})
\end{equation}

Stop rate can be expressed as the probability of risk above the stop threshold $t$:

\begin{equation}
\begin{aligned}
P(Z > t) = &P(X > g^{-1} (t) | Y = 1) P (Y = 1) + \\
        &P(X > g^{-1} (t) | Y = 0) P (Y = 0) 
\end{aligned}
\end{equation}

Let $H$ be the hit count.
\begin{equation}
H \sim Binomial(\text{stop count}, \text{hit rate})
\end{equation}

Hit rate can be expressed as the expectation of the risk above the stop threshold $t$:

\begin{equation}
E[Z|Z > t] = \frac{P(X > g^{-1} (t ) | Y = 1) P (Y = 1)}{P(Z>t)}
\end{equation}

\textit{Step 2: Across races, we fit the stop count per police region}:

Let $\mathbf{N}$ be a vector that contains the count of stops for races.
Let $N$ to be the sum of $\mathbf{N}$, the total count of stops between races in one precinct. 
Let $\mathbf{\theta}$ be a vector that contains the stop probabilities for races.
\begin{equation}
\mathbf{N} \sim Multinomial(N, \mathbf{\theta})
\end{equation}

The stop probability can be estimated from the stop rate and the US Census racial proportion, denoted as $c$. 

For example, the stop probability for the white race, $\theta_w$, follows:
\begin{equation}
    \theta_w  \propto P(Z_w > t _{w}){c _w}
\end{equation}
$c_w$ is the racial proportion of whites.
The white stop threshold $t_w$ and white stop rate $P(Z_w > t_w)$ are estimated in Step 1.

Incorporating the assumptions in the fast threshold test model is simple.
For Assumption 1,  
We adapt the model in following steps:
We repeat \textit{step} 1 and 2 for the stop and hit counts for daylight and darkness data. 
The stop thresholds for daylight stop and hit data and darkness stop and hit data are estimated in each run.
But they share the parameters in function $g$ in step 1 so the risk distribution remains one per race regardless of daylight and darkness.

For Assumption 2,
we use an additional constraint to ensure threshold at darkness is always smaller than the threshold at daylight.
First, the daylight threshold is sampled, same as before, from a standard normal distribution.
\begin{equation}
    t_{daylight} \leftarrow normal(0,1)
\end{equation}
Then, a parameter $t_{decay}$ is sampled as the decay from the daylight to darkness, also from a standard normal distribution.
We set the lower bound for  $t_{decay}$  as $0$.
\begin{equation}
    t_{decay} \leftarrow normal(0,1), \  
    t_{decay} \geq 0
\end{equation}
Then we set the threshold at darkness at the same race and location~(e.g., precinct),
\begin{equation}
    t_{darkenss} = t_{daylight} - t_{decay}
\end{equation}
All thresholds are then transformed to stay between $0$ and $1$ using a normal linkage function.

\begin{figure}
\vspace{.3in}
    \centering
    \includegraphics[width=0.80\columnwidth]{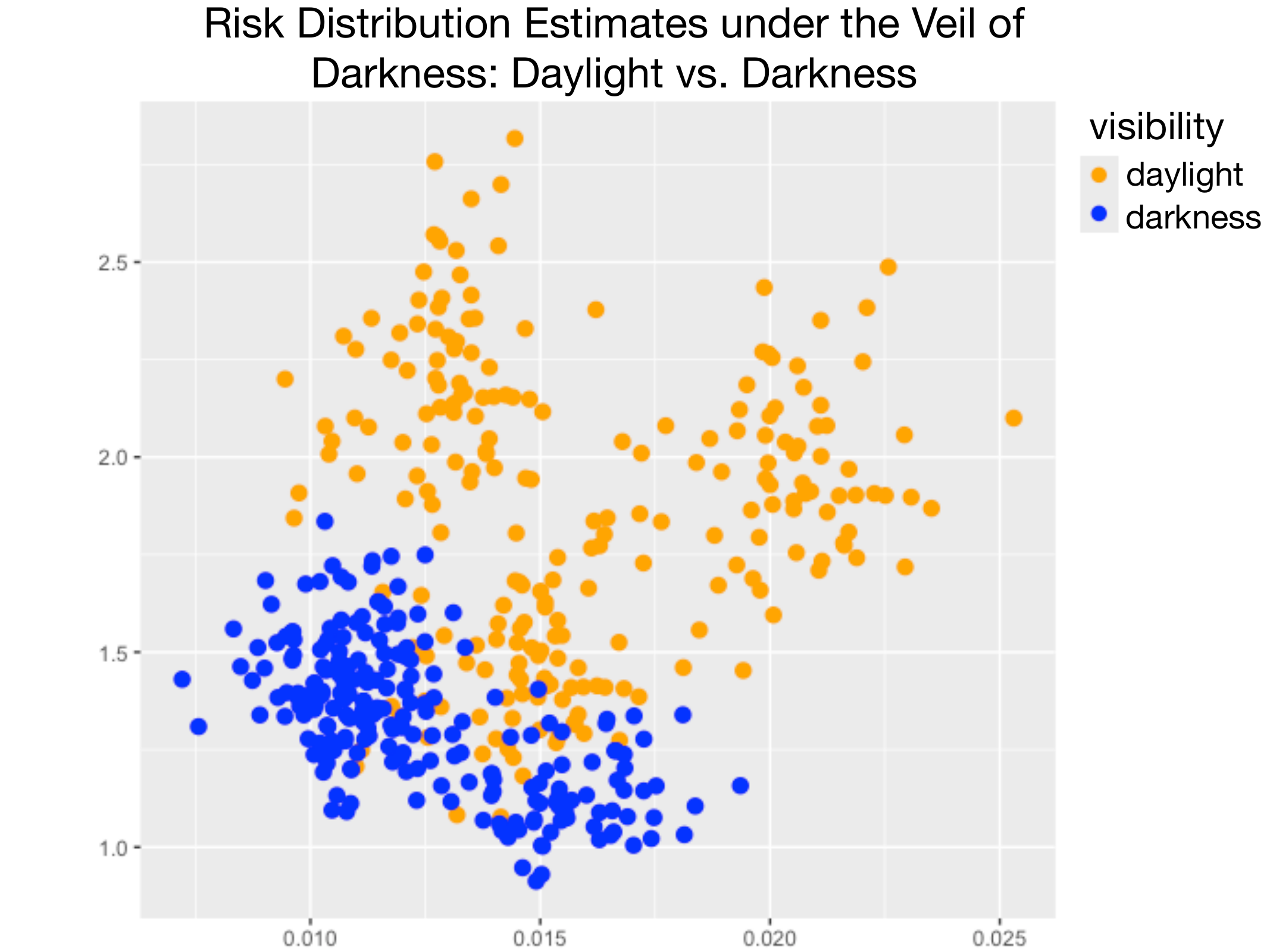}
    \caption{\textbf{Issue of threshold test.} 
    Estimated risk distribution parameters projected in 2D scatter plot. The vast difference between the orange and blue scatter plots indicates that the risk distributions between daylight and darkness are estimated to be very different, which does not align with the VoD assumption for distribution invariance per racial group.
    }
    \label{fig:exp1issues}
\end{figure}

\section{APPLICATION: NYC STOP-AND-FRISK DATA}
\label{sec:vod_data_prep}
We apply CCTT~(described in Section~\ref{sec:cctt}) on New York City stop-and-frisk records from 2008--2012. The dataset~\citep{pierson2018fast} contains 723{,}553 stops flagged as suspected contraband possession, along with 2010 Census estimates of racial composition by precinct. 
The distribution priors used in these threshold tests estimate separate risk distributions for each race within each precinct, capturing precinct-level differences in policing practices.
Using the same dataset and distribution priors as the basis of our experiments allows us to validate limitations, assess consistency, and directly compare our results with prior work. Notably, the official NYC stop-and-frisk portal provides data spans from 2003 to recent years. In more recent records, the rank of the police officer is also included.

\par \noindent \textbf{Data  and Curation}
We further create the curated data  $\mathcal{D}_K$ for VoD.
Since VoD is an effect that police stopping bias is reduced under low visibility~(i.e., darkness).
This implies that we want to include a stop event that may occur in either darkness or daylight in a given analysis scope~(e.g., time-span and geo-boundary). 
In other words, an event's counterfactual event \textit{has a chance to occur}. 
Such a counterfactual event only differs from the original event in visibility~(i.e., darkness or daylight).
For a time span of February 2018, and the geo-boundary of NYC, a stop at 8 PM local time would be impossible to occur in the daylight, unfit for understanding the veil of darkness on police stopping.
However, if the time span changes to the year 2018, such a stop can be included in the VoD analysis since 8 PM in May can be at daylight.
Such nuanced data curation based on analysis scope is crucial to the rigorousness of downstream statistical testing.
To reflect the VoD data condition,
We provide the curated VoD data~(146,491 records) \footnote{Code to reproduce the experiments is available online. \url{https://github.com/junyuan-ai/cctt-threshold-test}} for promoting reproducibility and transparency. The VoD data $\mathcal{D}_K$ is about 20 percent of the total $\mathcal{D}$.
We follow the tutorial of a recent work~\citep{knode2024pulling} for the data curation process. 
Here are the four steps:
1) We split the data by year.
2) For each year, we identify the stops during intertwilight.
3) For the stop in the intertwilight, we define stops between sunset and dusk as ambiguous stops, stops before sunset as daylight stops, and after dusk as darkness stops.
4) We exclude the ambiguous stops and aggregate the remaining stops across the year from $2008$ to $2012$.
The data process took about 5 hours on a 32 GB RAM, Intel MacBook Pro.
Note that one could choose not to split the data by year but set the time-span from the first day of $2008$ to the last day of $2012$, which will likely increase the record count of VoD data. As the possible range of the sunset and dawn time is wider from a longer time span than any sub-time span.
However, larger time-span increases the computational cost.

\par \noindent \textbf{Violation of Standard Threshold Test.}
\label{sec:dist_violate}
We apply the threshold test on both daylight and darkness data. The estimated group-level risk distributions differ sharply rather than overlaying between visibility conditions (Fig.~\ref{fig:exp1issues}), violating the VoD assumption that visibility blurs group signals without altering underlying risk distributions.

%% file: sections/simulationStudy.tex
We first validate the proposed cross-context VoD assumptions under controlled simulation. Then we apply them to more complex real-world policing data and adapt models from prior work.

\begin{figure*}
\vspace{.3in}
    \centering
    \includegraphics[width=0.85\textwidth]{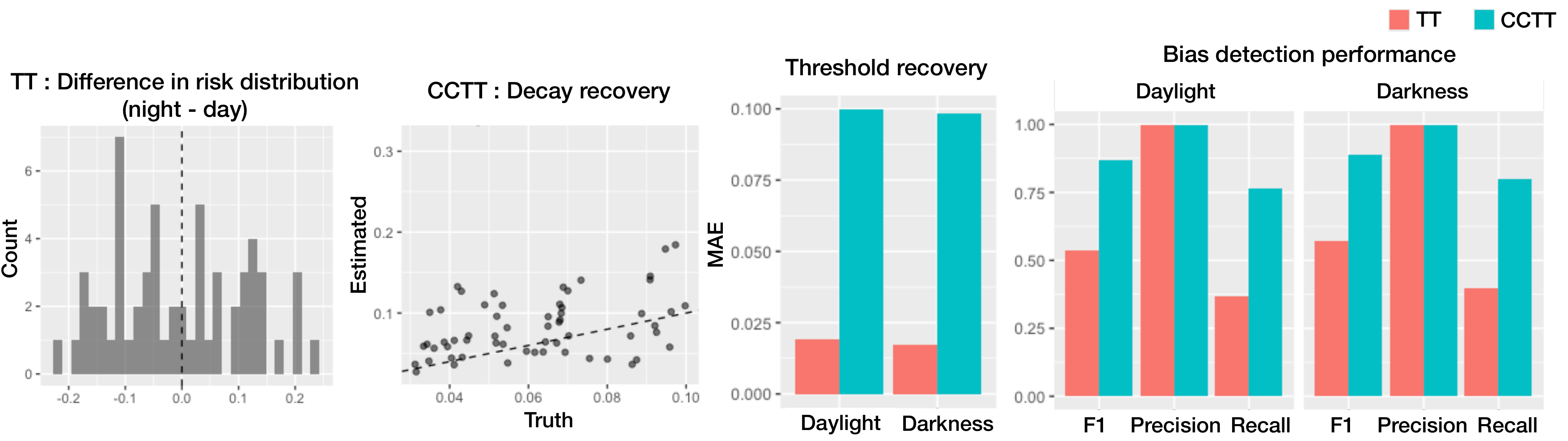}
    \caption{\textbf{Simulation study outcome.} From left to right,
    1) TT generates different day night risk distribution.
    2) CCTT estimates the decay is positively correlated with true decay.
    3) CCTT estimated the ground truth thresholds worse than TT. However, TT appears to overfit due to its flexibility in fitting separate distributions.
    4) The F1 and recall scores shows that CCTT is better at detecting the true bias than TT.
    }
    \label{fig:simulation}
\end{figure*}

\subsection{Simulation Study}
For simulation, we introduce the simplified CCTT.
Each group-precinct pair is assigned a single latent risk distribution,

\begin{equation}
    Z~{g, p} \sim \mathcal{N} (\mu_{g,p}, \sigma^2),
\end{equation}

Shared across lighting condition~(i.e., day and night).
Threshold follows 

\begin{equation}
    t_{g, p}^{\text{day}} \sim \mathcal{N} (\mu_{g,p}, 1)
\end{equation}

\begin{equation}
    t_{g, p}^{\text{night}} = t_{g, p}^{\text{day}}  - \delta_{g, p} ,\ \  \delta_{g, p} \geq 0  
\label{eqn:cctt_daynightConstraint}
\end{equation}

This simplified specification, including the use of a Normal distribution for the risk distribution, is intentional to isolate the roles of distributional invariance and monotone decay. To highlight the impact of applying VoD constraints on existing work, later sections adopt a Logit-Normal Mixture model instead of the Normal distribution, as it is more commonly used in prior threshold-based work~\citep{pierson2018fast, pierson2020large}.
Here, we assess identifiability and bias detection performance on simulated data.

For the context $c \in \{\text{day}, \text{night}\}$.
Let $N_{g,p}^c$ denotes the number of people in a racial group $g$ and police precinct $p$ at the context $c$.
The stop count is 

\begin{equation}
    S_{g,p}^c \sim Binomial(N_{g,p}^c, P(Z_{g,p} > t_{g,p}^c))
\label{eqn:cctt_stopDraw}
\end{equation}

The hit count is 
\begin{equation}
    H_{g,p}^c \sim Binomial(S_{g,p}^c, E[(Z_{g,p} > t_{g,p}^c])
\label{eqn:cctt_hitDraw}
\end{equation}

We report the following priors used for this simulation study.
$Z_{g,p} \sim \mathcal{N}(0, 1.5^2)$
$t_{g,p}^{\text{day}} \sim \mathcal{N}(0, 2^2)$
$\delta_{g,p} \sim HalfNormal(0.2, 0.4^2)$

A comparable standard threshold test~(TT) does not impose conditions of shared risk distribution or threshold decay.
For TT, we define and estimate the risk distributions $Z_{g,p}^c$ and thresholds $t_{g,p}^c$ separately for each context.
We omit the cross-context constraint (Equation~\ref{eqn:cctt_daynightConstraint}), while keeping the remaining steps (Equations~\ref{eqn:cctt_stopDraw} and~\ref{eqn:cctt_hitDraw}) unchanged.

\subsubsection{Generating Ground Truth}
We simulate 50 precincts for two racial groups (e.g., White and Black).  
Risk signals for class $Y \in \{\text{guilty}=1, \text{innocent}=0\}$ are generated as
\[
X \mid Y=0 \sim \mathcal{N}(0,1), \quad 
X \mid Y=1 \sim \mathcal{N}(\mu_1,1), \ 
\]
$\mu_1 \in [1.5, 2.5] $.
This construction ensures a separation between innocent and guilty individuals while producing realistic hit rates (10\%–20\%) when thresholds are applied.  

Daytime thresholds are assigned at the precinct level, with a race-specific gap:
\begin{equation}
    t_{g=\text{white},p}^{\text{day}} \sim U(0.35, 0.55),
\end{equation}
\begin{equation}
    t_{g=\text{black},p}^{\text{day}} = t_{g=\text{white},p}^{\text{day}} \pm \Delta_p, \quad \Delta_p \sim U(0.05, 0.15).
\end{equation}

Nighttime thresholds are defined by applying a non-negative decay to the daytime thresholds:
\begin{equation}
    t_{g,p}^{\text{night}} = t_{g,p}^{\text{day}} - \delta_{g,p}, \quad \delta_{g,p} \sim U(0.03, 0.10).
\end{equation}

The ground-truth population sizes $N_{g,p}^{\text{day}}$ and $N_{g,p}^{\text{night}}$ are drawn independently from a Poisson distribution with mean $2000$.  
For each individual, if $z > t$ the model records a stop $S$, and if additionally $Y=1$, a hit $H$.  

This setup guarantees that risk distributions are invariant across day and night while thresholds monotonically decrease in darkness.  
It also embeds true racial disparities through precinct-level gaps, providing a benchmark for evaluating whether the threshold test (TT) and our cross-context threshold test (CCTT) can correctly recover model parameters and detect bias.

\subsubsection{Results}
The simulation took approximately 5 minutes on a 32 GB RAM Intel MacBook Pro using R. The results are shown in Fig.~\ref{fig:simulation}.
Fig.~\ref{fig:simulation}.1 shows that the TT estimated the day and night risk distributions differ significantly, measured by the difference of $\mu$ between day and night. 
The differences are arbitrarily positive or negative but the actual difference should be negative.
Fig.~\ref{fig:simulation}.2 shows that the CCTT's estimation of threshold decay is positively correlated to the true decay.
Fig.~\ref{fig:simulation}.3 shows that the TT recovers thresholds with a lower Mean Absolute Error~(MAE), better than CCTT.
But considering Fig.~\ref{fig:simulation}.1, this indicates TT is over-fitted due the flexibility in two risk distributions. So relying on MAE is misleading in determining model performance.
Fig.~\ref{fig:simulation}.3 shows that actual performance in bias detection.
TT only detects bias when it’s absolutely sure~(i.e., high precision), so it misses many true biased precincts. 
CCTT has larger recall and overall higher F1, indicating better performance as bias detection method.

%% file: sections/caseStudy.tex
\begin{figure*}
\vspace{.3in}
    \centering
    \includegraphics[width=0.85\textwidth]{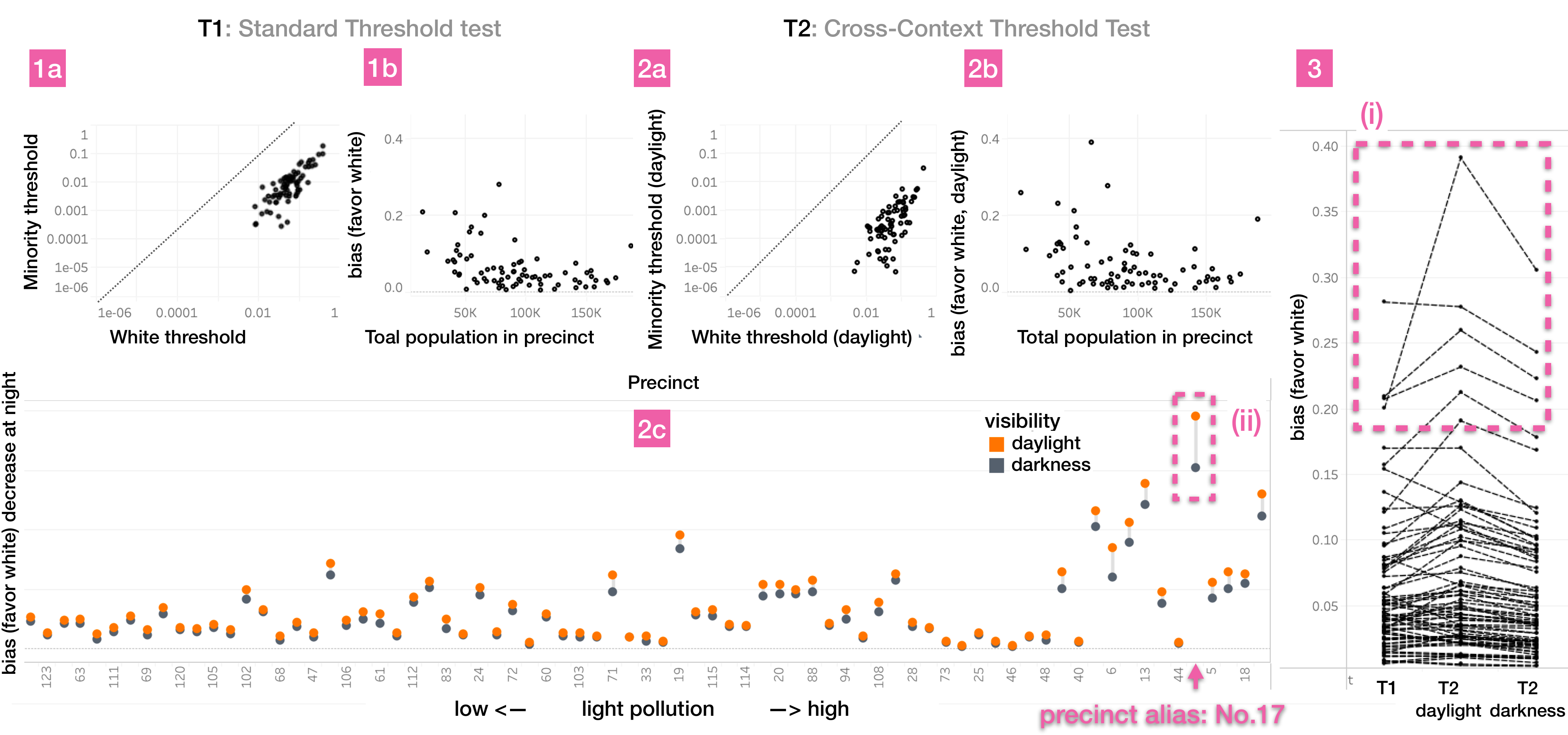}
    \caption{
    \textbf{Interactive audit dashboard for veil-of-darkness analysis.}
Panels show: threshold plots (1a,2a), bias plots (1b, 2b), contrastive bias plot (2c), and consensus plot (3), comparing bias judgment based on thresholds derived from the original TT and the CCTT.
Plots correspond to:
1a, 1b: threshold test without context.
2a, 2b, 2c: threshold test under VoD context
3: summarizes precinct-level bias values from all three tests using parallel coordinates, enabling auditors to see which precinct jurisdictions are flagged under each method.
    }
    \label{fig: dashboard}
\end{figure*}

\subsection{Visual Audit Dashboard}

As practical auditing may involve numerous contexts, traditional static plots is insufficient for comparing the effect of assumptions and decisions.
We demonstrate the application of our cross-context threshold test framework in a flexible and extendable visual dashboard~(Fig.~\ref{fig: dashboard})  for New York City stop-and-frisk veil-of-darkness (VoD) analysis. 
The dashboard was developed to address the oversight challenge that human auditors must often compare results from multiple bias detection methods, interpret the magnitude and direction of disparities, and decide which jurisdictions warrant further investigation.
Prior work~\citep{pierson2018fast} uses a static threshold plot to show the pattern of thresholds between two different groups~(e.g., white and minority racial groups).
Threshold plots~(Fig.\ref{fig: dashboard}(1a, 2a)) directly encode the thresholds of two groups for the same police precinct, but the degree of bias can only be visually perceived as the distance from the dot to the diagonal line~(i.e., dots falling on the diagonal line indicates the equal thresholds the precincts use for stop decision).
The static threshold plot also does not support flexible exploration and iterative investigation of the context or the instances~(e.g., police precincts) under investigation, which highlight the importance of interactivity in an audit dashboard.
Note that the Fig.\ref{fig: dashboard}(1a) reproduces the Figure 4 (left) in prior work~\citep{pierson2018fast}, omitting the size difference of dots~(which encoded the population size).  
Fig.\ref{fig: dashboard}(2a) is the cross-context threshold plot with the exponential axes scaling same as Fig.\ref{fig: dashboard}(1a).  Comparing the figures~(Fig.\ref{fig: dashboard}(1a, 2a)) tells us that the threshold pattern and bias conclusion remains the same between the original threshold test and CCTT--the police favors the white group. 

We introduce three additional plots~(bias plot, contrastive bias plot, and consensus plot) to support direct encoding of bias and cross-context comparison. 
The bias plot~(Fig.\ref{fig: dashboard}(1b, 2b))  directly encodes the magnitude of disparity as the vertical distance from a fairness baseline.
The contrastive bias plot~(Fig.\ref{fig: dashboard}(2c)) focuses on the change in disparity between daylight and darkness~(i.e., the threshold decay). 
The consensus plot~(Fig.\ref{fig: dashboard}(3))  summarizes results from multiple bias test outcomes, allowing auditors to identify precincts flagged as biased under at least one case.

The dashboard supports interactive filtering and cross-highlighting across plots, enabling an auditor to, for example, select precincts with bias above a chosen threshold in one plot and immediately see their status in others. 
This allows auditors to discover precincts where disparities are both large and consistent across bias test methods~(Fig.\ref{fig: dashboard}(i)).

For the contrastive bias plot (Fig.\ref{fig: dashboard}(2c)), we bring in light pollution data to infer the effect of the veil-of-darkness (VoD) from underlying policing practices. We use the average masked composite from the VIIRS Nighttime Lights (VNL V2) dataset~\citep{elvidge2021annual}, which provides a stable measure of persistent anthropogenic light. 
Transient sources such as fires and boats are removed through masking, ensuring data reflect long-term urban light pollution. This dataset balances between spatial resolution ($\sim$500 m) and file size~($\sim$262 MB), and is sufficient for precinct-level aggregation in New York City.

In Fig.~\ref{fig: dashboard}(2c), the precinct at the far right corresponds to Times Square, which exhibits the highest light pollution. 
According to the VoD logic, policing outcomes should be fairer at night, since reduced visibility implies that officers rely on lower thresholds across racial groups. However, this does not mean that they are affected equivalently due to light pollution.
We formalize stop bias within an explanatory causal model (ECM)~\footnote{While we adopt ECM as a conceptual tool, we do not claim that our observational design fully identifies all causal effects (see \citet{banitz2022model} for caveats of model-derived explanation).}, which specifies bias as the causal effect of visibility and policing practice. The visibility affects the police practice, then police practice affects the stop bias, so changes in stop bias cannot be interpreted in isolation. 
In structural form,

\begin{tikzpicture}
    \node (1) at (0,0) {Visibility};
    \node (2) [right = of 1] {Police Practice};
    \node (3) [right = of 2] {Stop Bias};
    \path (1) edge (2);
    \path (2) edge (3);
\end{tikzpicture}
In precincts with high levels of light pollution, the difference between daylight and darkness is minimal, yet we observe substantial reductions in stop bias~(e.g., precinct alias No.17). Since visibility effects are unlikely to explain such drops, these reductions are more plausibly associated with changes in policing practices. 
This does not imply that policing in these contexts is unbiased or fair. A reduction in disparity under specific conditions should be interpreted only as evidence of practice shifts, not as the absence of bias~(e.g., precinct No.17 has large bias even after bias decrease at night).
In precincts with low levels of light pollution, where the transition from daylight to darkness substantially reduces visibility~(e.g., precinct on the far left of the Figure~\ref{fig: dashboard}(2c)), one would expect bias to diminish if officers relied less on racial cues. Yet our results show little or no reduction. This persistence suggests that policing practices may actively counteract the expected visibility effect, sustaining disparities even when race is less identifiable. 
Using ECM framing, we can disentangle environmental influences from institutional practices and prevent over-interpretation of statistical outcomes as straightforward evidence of fairness.
Our study does not claim full causal identification, but to align the cross-context threshold test framework with broader causal inference literature and clarify how observed disparities should be interpreted.

%% file: sections/Discussion.tex
\section{DISCUSSION}

While our study emphasized visibility as the primary environmental context shaping bias detection, the framework naturally extends to other contextual dimensions within policing and to domains beyond law enforcement.
Within policing, institutional roles create additional layers of decision-making. Patrol officers, supervisors, and specialized units differ in authority, oversight, and exposure. Similar to the veil-of-darkness (VoD) setting, the risk distribution of racial groups should remain invariant across these organizational contexts. Differences in estimated thresholds therefore reflect institutional practices rather than population risk heterogeneity. Moreover, environmental and organizational factors can intersect: for example, supervisors may apply different daylight versus darkness thresholds than frontline officers. Such intersections enrich contextual auditing.

The CCTT framework extends beyond policing. In AI model auditing, visibility can be reinterpreted as the amount of context available in prompts: high-context prompts expose more features, while low-context prompts obscure them, analogous to visibility changes in policing. CCTT allows auditors to separate shifts in underlying ``risk distributions'' from shifts in decision thresholds, clarifying whether disparities stem from model mechanisms or input conditions. In medical imaging, differing image resolutions create challenges: low-quality scans may obscure signals from similar disease~(e.g., pneumonia and lung cancer), shifting diagnostic thresholds while disease prevalence remains constant. By curating datasets across imaging conditions and holding latent distributions fixed, threshold analysis can reveal systematic decision shifts attributable to degraded visibility.
These examples show that the CCTT framework is flexible and generalizable.
Furthermore, embedding context directly into threshold tests yields more interpretable and actionable evidence. 

CCTT has several limitations.
First, threshold tests require specifying a parametric form for the risk distribution.
Choices such as Normal, Beta, or logit-normal mixtures may still fail to capture the true underlying distributions.
Second, while CCTT runs in comparable time to the fast threshold test, it still requires minutes of MCMC sampling in R.
This constrains flexibility and reduces the ability to redefine contexts interactively in an auditing dashboard.

%% file: sections/conclusion.tex
\section{CONCLUSION}

We introduce the Cross-Context Threshold Test (CCTT), a framework for adapting threshold tests to settings where environmental or institutional context influences decision-making.  
Standard threshold tests, when applied separately to daylight and darkness, violate the assumptions of the veil-of-darkness (VoD) framework by allowing risk distributions to drift arbitrarily and thresholds to fluctuate inconsistently.  
CCTT addresses these issues by explicitly enforcing distributional invariance across contexts and monotonic threshold decay, aligning statistical tests with the counterfactual logic of VoD.  
Through simulation studies, we show that CCTT yields higher recall and F1 scores while maintaining perfect precision in detecting biased precincts compared to the standard threshold test.  
Applied to New York City stop-and-frisk data, CCTT uncovered additional instances of racial disparities that are overlooked by existing approaches.  
We also provide a visualization tool to illustrate threshold shifts and cross-context comparisons, highlighting how these methodological advances can be interpreted for oversight. 
While this paper focused on visibility as an environmental context, the framework extends naturally to organizational settings (e.g., officer rank or unit) and to other domains where context alters decision thresholds without changing underlying risk distributions.  
Examples include medical imaging under varying resolutions and AI auditing under different prompt conditions.  
By embedding contextual assumptions directly into threshold tests, CCTT advances both the methodological rigor and the practical utility of fairness audits.